\documentclass[runningheads]{llncs}

\PassOptionsToPackage{table,dvipsnames}{xcolor}

\usepackage{eccv}

\usepackage{eccvabbrv}
\usepackage{graphicx}
\usepackage{booktabs}
\usepackage[accsupp]{axessibility}

\usepackage{multirow}

\colorlet{mediumgray}{gray!60}
\colorlet{lightergray}{gray!20}
\definecolor{lightred}{RGB}{255,200,200}
\definecolor{lightgreen}{RGB}{200,255,200}
\definecolor{lightyellow}{RGB}{255,255,200}
\definecolor{lightgray}{RGB}{200,200,200}

\usepackage{algorithm}
\usepackage{algorithmic}
\usepackage{placeins}   
\usepackage{microtype}
\usepackage{hyperref}

\usepackage{orcidlink}

\makeatletter
\newcommand{\fakesupplabel}[2]{%
  \@namedef{r@#1}{{#2}{#2}{}{}{}}%
}
\makeatother

\fakesupplabel{sec:rationale}{A}
\fakesupplabel{sec:supp_fmcw}{A.1}
\fakesupplabel{sec:supp_iq}{A.2}
\fakesupplabel{sec:tdm_mimo_imaging}{A.3}
\fakesupplabel{sec:supp_mc_derivation}{A.4}
\fakesupplabel{sec:supp_range_angle_resolution}{A.5}
\fakesupplabel{sec:supp_polar_to_cartesian}{A.6}
\fakesupplabel{sec:supp_implementation}{B}
\fakesupplabel{sec:supp_preprocessing}{B.1}
\fakesupplabel{sec:supp_alignment}{B.2}
\fakesupplabel{sec:phase_sensitivity_loss_landscape}{B.3}
\fakesupplabel{sec:supp_training}{B.4}
\fakesupplabel{sec:supp_inference}{B.5}
\fakesupplabel{sec:supp_postprocess}{B.6}
\fakesupplabel{sec:supp_qualitative}{C}
\fakesupplabel{sec:qual_materials}{C.1}
\fakesupplabel{sec:qual_normals}{C.2}
\fakesupplabel{sec:qual_beam}{C.3}
\fakesupplabel{sec:ablation}{C.4}
\fakesupplabel{sec:runtime}{C.5}
\fakesupplabel{sec:limitations}{D}
\fakesupplabel{sec:supp_failures}{D.7}

\begin{document}

\title{mmIR: Frequency-Space Inverse Rendering for 3D Millimeter-Wave Radar ADC Synthesis}

\titlerunning{mmIR: Inverse Rendering for mmWave Radar ADC Synthesis}

\author{
  Adnan Armouti\inst{1}\orcidlink{0000-0002-4448-3089} \and
  Yixuan Gao\inst{1}\orcidlink{0000-0003-1778-3104} \and
  Rajalakshmi Nandakumar\inst{1}\orcidlink{0000-0002-1601-148X}
}

\authorrunning{A.~Armouti et al.}

\institute{Cornell Tech, New York, NY, USA\\
\email{\{aa2546,yg478,rn283\}@cornell.edu}}

\maketitle

\begin{abstract}
High-resolution 3D radar data is scarce. Commodity mmWave sensors use small antenna arrays that limit angular resolution to several degrees, and existing datasets provide only 2D range--azimuth maps or sparse point clouds rather than raw analog-to-digital converter (ADC) signals. Hardware scaling is expensive, synthetic-aperture scanning is impractical at fleet scale, and learned synthesis methods are bottlenecked by the very data shortage they aim to address.
We present mmIR, an open-source differentiable frequency-modulated continuous-wave (FMCW) radar inverse renderer that fits a physics-based forward model to real captures and re-renders from dense virtual apertures to synthesize high-resolution 3D radar data. Because radar resolution is too coarse to recover geometry directly, mmIR performs LiDAR-assisted inverse rendering: using LiDAR-derived meshes as a geometric scaffold, mmIR optimizes per-vertex International Telecommunication Union (ITU) physics materials, vertex normals, and antenna beam patterns through end-to-end automatic differentiation of a phase-coherent multiple-input multiple-output (MIMO) forward model with multi-bounce propagation, polarization, and free-space diffraction.
On seven outdoor and six indoor ColoRadar scenes, mmIR achieves 0.914 mean Pearson correlation on range--azimuth maps versus 0.307 for Sionna-RT. Scenes trained on a cascaded imaging radar transfer to a co-located single-chip radar without re-training (0.554 correlation), and dense virtual arrays (100$\times$100 elements) produce single-frame 3D occupancy validated against LiDAR.
Project page: \url{https://mmwave-inverse-rendering.github.io/}.

\end{abstract}

\section{Introduction}

\begin{figure}[t]
  \centering
  \includegraphics[width=\linewidth]{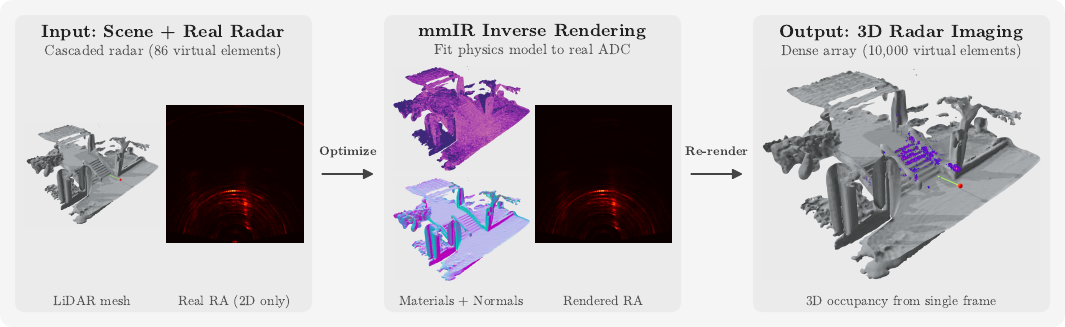}
  \caption{Given a LiDAR mesh and GT real cascaded-radar RA map (left), mmIR fits per-vertex materials, normals, and beam patterns (center), then re-renders a dense 100$\times$100 virtual array for single-frame 3D occupancy (right).}
  \label{fig:teaser}
\end{figure}

Millimeter-wave (mmWave) radar penetrates fog, rain, and airborne dust~\cite{9157693}, measures Doppler velocity, and returns strong echoes from metals, yet it remains underutilized for 3D scene understanding. Commercial single-chip radars have only tens of antennas, limiting angular resolution to several degrees~\cite{9048939}, and their axis-biased layouts yield 2D range--azimuth (RA) slices rather than full 3D range--azimuth--elevation (RAE) volumes. Public datasets further limit access by providing only post-processed outputs (RA maps or sparse point clouds) rather than raw ADC~\cite{9196884, Caesar_2020_CVPR}. The result is a shortage of high-resolution, 3D, raw FMCW ADC data, the fundamental bottleneck for learning-based radar perception. Detection, segmentation, occupancy prediction, and super-resolution all require training signal that current sensors and datasets cannot supply, and addressing this gap would enable these applications.

Existing approaches each fall short. Hardware scaling~\cite{9127853} increases cost without addressing data diversity. SAR~\cite{7759633, s23135979, s17061419} requires mechanical scanning impractical at fleet scale. Neural and Gaussian synthesis methods~\cite{Mostajabi_2020_CVPR_Workshops, 10.1145/3641519.3657510, Huang_2024_CVPR, takawale2025spinrneuralvolumetricreconstruction, 11444848} are bottlenecked by the very data scarcity they aim to address. Models trained on low-resolution 2D RA maps cannot recover 3D structure that was never observed.

Physics-based ray tracing (SBR) for mmWave is well established~\cite{5728244, 10465179}, but existing pipelines are forward-only, requiring geometry and materials to be specified \emph{a priori}, with no mechanism to calibrate against real measurements. Differentiable rendering has closed this gap for RGB~\cite{10.1145/3272127.3275109, 9879300}, LiDAR~\cite{kato2020differentiablerenderingsurvey}, and acoustics~\cite{10.1145/3641519.3657493}, and emerging RF methods take initial steps. Sionna-RT~\cite{10465179} exposes gradients for antenna patterns and scattering coefficients but not through ray tracing itself (no geometry or normal gradients) and does not model wideband FMCW signals. Hofmann et al.~\cite{10892639} estimate SFCW materials but assume known geometry with single-bounce propagation, and InverTwin~\cite{chen2025invertwinsolvinginverseproblems} addresses multipath gradient pathologies but validates on synthetic data only. What is missing is an end-to-end differentiable FMCW renderer that can learn scene parameters from real radar captures without requiring ground-truth materials.

We present mmIR, an open-source differentiable FMCW radar inverse renderer that fits a physics-based forward model to real captures and re-renders from dense virtual apertures to synthesize high-resolution 3D radar data. Radar resolution is too coarse to reconstruct geometry directly, so mmIR adopts a LiDAR-assisted formulation, using LiDAR-derived meshes~\cite{1281957.1281965} as a geometric scaffold, a practical choice given the far greater availability and resolution of LiDAR data. Inverse rendering then recovers the electromagnetic properties. mmIR optimizes per-vertex ITU physics materials (permittivity, roughness, correlation length, blend factor, slab thickness), vertex normals, and antenna beam patterns through end-to-end automatic differentiation of a phase-coherent MIMO forward model with multi-bounce propagation, polarization-aware bidirectional scattering distribution functions (BSDFs), specular manifold sampling~\cite{10.1145/3386569.3392408}, and free-space diffraction~\cite{10.1145/3658166}. The pipeline also supports differentiable geometry and 6-DoF pose gradients with projective boundary sampling~\cite{10.1145/3618385}, though we keep these fixed in our experiments because LiDAR-radar alignment preprocessing (Sec.~\ref{sec:supp_alignment} in the supplement) already provides accurate geometry and pose. After fitting, the optimized scene is frozen and re-rendered from arbitrarily large virtual apertures to synthesize dense, physics-consistent raw ADC (Fig.~\ref{fig:teaser}), enabling downstream applications such as 3D occupancy detection, radar super-resolution, and synthetic training data generation that would otherwise require prohibitively large physical arrays.

On seven outdoor and six indoor ColoRadar~\cite{doi:10.1177/02783649211068535} scenes, mmIR achieves 0.914 mean Pearson RA correlation versus 0.307 for Sionna-RT. We demonstrate cross-sensor transfer. Scenes trained on a cascaded imaging radar (12\,transmit (Tx)$\times$16\,receive (Rx) antennas spanning 86 uniformly sampled virtual elements along azimuth) produce valid ADC for a co-located single-chip radar (3Tx$\times$4Rx, 12 virtual elements total spanning 8 uniformly sampled virtual elements along azimuth) without re-training (0.554 correlation), providing evidence that the optimized scenes capture transferable physics rather than sensor-specific patterns. We further show single-frame 3D occupancy detection from dense virtual arrays (100Tx$\times$100Rx) validated against LiDAR. mmIR's scene fitting takes ${\sim}11$\,min (500 iterations, $1.32$\,s/iter) versus $3.28$\,s/iter for Sionna-RT (Sec.~\ref{sec:runtime}), and a dense-array frame renders in $6.4$\,s on a single NVIDIA RTX 4090. Our contributions are:

\begin{enumerate}
\item To the best of our knowledge, the first open-source, end-to-end differentiable inverse renderer for wideband FMCW radar, enabling joint optimization of per-vertex materials, normals, and antenna patterns directly from real radar captures, with support for multi-bounce propagation, specular manifold sampling, polarization-aware BSDFs, free-space diffraction, and differentiable geometry and pose gradients with projective boundary sampling.
\item A virtual-aperture re-rendering framework that synthesizes high-resolution 3D raw ADC from fitted scenes via arbitrarily large synthetic arrays, enabling radar resolution lifting from 2D RA to 3D RAE on the provided LiDAR scaffold, and 3D occupancy detection.
\item Cross-sensor transfer validation on real COTS radar, demonstrating that scenes optimized on one sensor generalize to a different configuration without re-training.
\end{enumerate}

\subsection{Scope}
\label{sec:scope}

mmIR targets high-quality, physics-consistent raw-ADC synthesis by fitting a differentiable forward model on LiDAR-derived geometry to real commercial off-the-shelf (COTS) radar measurements. We envision practitioners using optimized scenes to build large-scale datasets for training downstream perception models such as detection, segmentation, and occupancy prediction, without additional data collection. We show cross-sensor transfer and single-frame 3D occupancy from dense virtual arrays (Sec.~\ref{sec:3d_occupancy}), comparing only against open-source physics-based ray-tracing renderers (Table~\ref{tab:polarization_comparison}) for a reproducible, like-for-like evaluation. We also note that single-frame radar inverse rendering is severely ill-posed. Our experiments are run on a single view with a single frame, which suffers from sparse returns, elevation integrated out by 2D RA, and resolution far coarser than LiDAR. Therefore, we adopt a mesh-based geometric prior following Sionna-RT~\cite{10465179}, but validate on meshes obtained from real LiDAR data rather than synthetic CAD, and fit to real radar measurements instead of synthetic channel responses. We do not claim shape-from-radar, an open problem under single-frame supervision that is out of scope. Finally, we assume static captures, leaving Doppler (ego-motion) and dynamic scenes to future work.

\begin{table}[t]
\centering
\caption{Feature comparison of physics-based differentiable radar renderers. Signal: ADC\,=\,raw FMCW, CIR\,=\,channel impulse response, IF\,=\,intermediate-frequency phasors. Multi-b.\,=\,multi-bounce; Diffr.\,=\,edge diffraction; Polar.\,=\,polarization; Geom.\,=\,differentiable geometry; Norm.\,=\,differentiable normals; Mat.\,=\,physics material model; Per-vtx\,=\,per-vertex materials; Bound.\,$\nabla$\,=\,boundary gradients; Real\,=\,validated on real data. Checkmarks denote supported pipeline capabilities; see Sec.~\ref{sec:supp_training} in the supplement for which parameters are active in our experiments.}
\label{tab:polarization_comparison}
\small
\setlength{\tabcolsep}{3pt}
\renewcommand{\arraystretch}{1.1}
\resizebox{\linewidth}{!}{
\begin{tabular}{|l|c|c|c|c|c|c|c|c|c|c|c|c|c|c|}
\hline
\rowcolor{mediumgray}
\textbf{Method} &
\textbf{Open} &
\textbf{Repr.} &
\textbf{Signal} &
\textbf{Multi-b.} &
\textbf{Diffr.} &
\textbf{Polar.} &
\textbf{Geom.} &
\textbf{Norm.} &
\textbf{Mat.} &
\textbf{Pose} &
\textbf{Beam} &
\textbf{Per-vtx} &
\textbf{Bound. $\nabla$} &
\textbf{Real} \\
\hline
\cellcolor{lightergray}Sionna-RT~\cite{10465179} &
\cellcolor{lightgreen}$\checkmark$ &
\cellcolor{lightgreen}Mesh &
\cellcolor{lightyellow}CIR &
\cellcolor{lightgreen}$\checkmark$ &
\cellcolor{lightgreen}$\checkmark$ &
\cellcolor{lightgreen}$\checkmark$ &
\cellcolor{lightred}-- &
\cellcolor{lightred}-- &
\cellcolor{lightyellow}No BRDF &
\cellcolor{lightgreen}$\checkmark$ &
\cellcolor{lightred}-- &
\cellcolor{lightred}-- &
\cellcolor{lightred}-- &
\cellcolor{lightred}-- \\
\hline
\cellcolor{lightergray}SFCW Inv.~\cite{10892639} &
\cellcolor{lightgreen}$\checkmark$ &
\cellcolor{lightgreen}Mesh &
\cellcolor{lightyellow}IF &
\cellcolor{lightred}-- &
\cellcolor{lightred}-- &
\cellcolor{lightgreen}$\checkmark$ &
\cellcolor{lightred}-- &
\cellcolor{lightyellow}Shading &
\cellcolor{lightgreen}$\checkmark$ &
\cellcolor{lightyellow}Pos. &
\cellcolor{lightred}-- &
\cellcolor{lightgreen}$\checkmark$ &
\cellcolor{lightred}-- &
\cellcolor{lightgreen}$\checkmark$ \\
\hline
\cellcolor{lightergray}InverTwin~\cite{chen2025invertwinsolvinginverseproblems} &
\cellcolor{lightred}-- &
\cellcolor{lightgreen}Mesh &
\cellcolor{lightyellow}CIR &
\cellcolor{lightgreen}$\checkmark$ &
\cellcolor{lightgreen}$\checkmark$ &
\cellcolor{lightred}-- &
\cellcolor{lightgreen}$\checkmark$ &
\cellcolor{lightred}-- &
\cellcolor{lightyellow}No BRDF &
\cellcolor{lightgreen}$\checkmark$ &
\cellcolor{lightred}-- &
\cellcolor{lightred}-- &
\cellcolor{lightgreen}$\checkmark$ &
\cellcolor{lightgreen}$\checkmark$ \\
\hline
\cellcolor{lightergray}mmIR (Ours) &
\cellcolor{lightgreen}$\checkmark$ &
\cellcolor{lightgreen}Mesh &
\cellcolor{lightgreen}ADC &
\cellcolor{lightgreen}$\checkmark$ &
\cellcolor{lightgreen}$\checkmark$ &
\cellcolor{lightgreen}$\checkmark$ &
\cellcolor{lightgreen}$\checkmark$ &
\cellcolor{lightgreen}$\checkmark$ &
\cellcolor{lightgreen}$\checkmark$ &
\cellcolor{lightgreen}$\checkmark$ &
\cellcolor{lightgreen}$\checkmark$ &
\cellcolor{lightgreen}$\checkmark$ &
\cellcolor{lightgreen}$\checkmark$ &
\cellcolor{lightgreen}$\checkmark$ \\
\hline
\end{tabular}
}
\end{table}

\section{Related Works}

\textbf{Differentiable rendering beyond optics.}
Differentiable rendering couples explicit scene parameterizations with gradient-based optimization to recover geometry, materials, and sensor parameters~\cite{10.1145/218380.218498, veach1998robust, 10.1145/1189762.1189764, 4409003}. Core advances include analytic inverse rendering and Monte Carlo (MC) estimator reparameterizations~\cite{10.1145/3272127.3275109, 10.1145/3355089.3356510, 10.1145/3414685.3417833}, as well as practical path tracing systems and neural/hybrid representations~\cite{10.1145/3503250, 10.1145/3550469.3555397, 9879300, 10.1145/3355089.3356498, 10.1145/3528223.3530099, 10.1145/3618353}. These methods operate almost exclusively in the optical regime. In RF, Sionna-RT~\cite{10465179} propagates gradients through field coefficients and CIR computation but not through path topology. Its path solver runs under gradient suspension and subsequently detaches all path geometry, so vertex positions and surface normals receive no gradients, restricting inverse rendering to quantities that enter only the CIR computation (antenna patterns, scattering coefficients, Tx/Rx positions). It also does not model wideband FMCW signals. Hofmann et al.~\cite{10892639} estimate SFCW materials but require known geometry, model only single-bounce propagation, and treat visibility as non-differentiable. InverTwin~\cite{chen2025invertwinsolvinginverseproblems} addresses multipath gradient pathologies via path-space differentiation but substitutes a Gaussian kernel surrogate for radar range processing and validates on synthetic data only. mmIR bridges this gap with end-to-end AD through ray tracing itself, enabling gradients to flow through all bounces back to geometry, materials, and sensor parameters directly from raw frequency-space FMCW ADC (Table~\ref{tab:polarization_comparison}).

\begin{figure}[t]
  \centering
  \includegraphics[width=\linewidth]{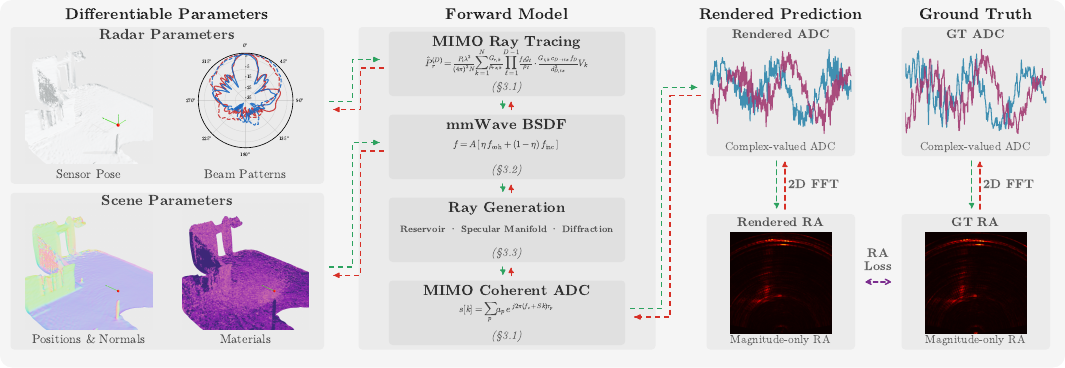}
  \caption{mmIR pipeline. Given a LiDAR mesh and sensor parameters, the differentiable model traces phase-coherent MIMO paths (reflection + diffraction) into raw FMCW ADC. End-to-end AD (ray tracing, BSDF, phasor accumulation, ADC) optimizes per-vertex materials, normals, and antenna beam patterns from a range--azimuth loss.}
  \label{fig:system}
\end{figure}

\textbf{Ray tracing for FMCW signal modeling.}
Radar simulation typically employs shooting-and-bouncing rays (SBR) to scale beyond full-wave solvers, producing path geometry and attenuation for large scenes and MIMO arrays~\cite{5728244}. Recent computational RF imaging further demonstrates multi-bounce physics for reconstruction on real mmWave hardware~\cite{dodds2025mitomillimeterwavedatasetsimulator}. However, these pipelines are forward-only. Scene and sensor parameters are fixed, precluding calibration against real captures. mmIR makes SBR-style FMCW rendering fully differentiable, exposing gradients w.r.t.\ mesh geometry, per-vertex materials, radar pose, and antenna patterns to close the sim-to-real gap.

\textbf{Neural and Gaussian methods for radar data synthesis.}
Neural fields have been adapted to radar across modalities, including SAR/ISAR shape reconstruction~\cite{lei2023sarnerfneuralradiancefields, 10423594}, FMCW occupancy and Doppler recovery~\cite{takawale2025spinrneuralvolumetricreconstruction, Huang_2024_CVPR}, frequency-space response synthesis~\cite{10.1145/3641519.3657510, Zhang_2026_CVPR}, and multi-modal fusion with RGB/LiDAR~\cite{Rafidashti_2025_CVPR, lu2024newrfdeeplearningframework}. RadarSplat~\cite{11444848} provides a Gaussian splatting~\cite{10.1145/3592433} alternative for range--azimuth rendering. These methods learn opaque, sensor-specific representations. DART~\cite{Huang_2024_CVPR} discards phase entirely (magnitude-only range--Doppler supervision), none expose physically interpretable material parameters, and none model multi-bounce propagation, polarization, or diffraction. Moreover, methods that render post-processed images directly must learn to reproduce FFT point-spread artifacts (sidelobes, windowing). By rendering raw ADC, mmIR applies identical FFT processing to both rendered and measured data, cleanly decoupling scene physics from signal processing. mmIR maintains an explicit mesh with per-vertex ITU materials and polarization-aware BSDFs, enabling joint scene--sensor calibration and re-rendering from arbitrary virtual apertures for 3D imaging.

\section{Methods}
\label{sec:methods}

We model radar sensing as a forward rendering problem (Fig.~\ref{fig:system}), tracing multi-bounce rays through a scene, evaluating surface interactions via a physically-based mmWave BSDF, and accumulating phase-coherent paths into raw FMCW ADC samples that can be directly compared against real captures. Sec.~\ref{sec:forward_model} derives the forward model from the bistatic radar equation through multi-bounce Monte Carlo estimation to ADC synthesis; Sec.~\ref{sec:bsdf} defines the mmWave BSDF governing surface interactions; Sec.~\ref{sec:sampling} describes ray generation strategies; and Sec.~\ref{sec:optimization} formulates the differentiable inverse rendering pipeline.

Our inputs are synchronized radar--LiDAR data from ColoRadar~\cite{doi:10.1177/02783649211068535}. LiDAR frames are aggregated into a triangle mesh via screened Poisson reconstruction~\cite{1281957.1281965}, and a GPU-accelerated coarse-to-fine alignment corrects residual sensor misregistration (Sec.~\ref{sec:supp_preprocessing}--\ref{sec:supp_alignment} in the supplement). FMCW signal and TDM-MIMO imaging derivations are in the supplement (Sec.~\ref{sec:rationale}--\ref{sec:tdm_mimo_imaging}).

\subsection{Forward Model}
\label{sec:forward_model}

Our forward model is grounded in the bistatic radar equation. For a single Tx--Rx pair observing a point target at surface point $\mathbf{x}$ with normal $\mathbf{n}$, at distances $d_t$, $d_r$ from transmitter and receiver, the received power $P_r$ is:
\begin{equation}
    P_r
    =
    \frac{P_t\, G_t\, G_r\, \lambda^2\, \sigma}{(4\pi)^3\, d_t^2\, d_r^2},
    \label{eq:radar_bistatic}
\end{equation}
where $P_t$ is transmit power, $G_t$, $G_r$ are directional antenna gains, $\lambda$ is wavelength, and $\sigma$ is the bistatic radar cross-section. Replacing the point-target RCS with a spatially varying BSDF $f(\boldsymbol{\omega}_t, \mathbf{x}, \boldsymbol{\omega}_r)$, where $\boldsymbol{\omega}_t$ and $\boldsymbol{\omega}_r$ are unit directions toward the transmitter and receiver, and converting from surface area element $dA_x$ to receiver solid angle via $d\omega_r = (c_r / d_r^2)\,dA_x$ with $c_r = \max(0, \mathbf{n} \cdot \boldsymbol{\omega}_r)$ (derivation in the supplement, Sec.~\ref{sec:supp_mc_derivation}), the total received power integrated over the receiver hemisphere $\Omega_{rx}$ becomes:
\begin{equation}
    P_r
    =
    \frac{P_t\,\lambda^2}{(4\pi)^2}
    \int_{\Omega_{rx}}
    \frac{
    G_t(\boldsymbol{\omega}_t)\,G_r(\boldsymbol{\omega}_r)\,
    f(\boldsymbol{\omega}_t, \mathbf{x}, \boldsymbol{\omega}_r)\,
    c_t\, V
    }{
    d_t^2
    }
    \, d\omega_r,
    \label{eq:radar_solid_angle}
\end{equation}
where $c_t = \max(0, \mathbf{n} \cdot \boldsymbol{\omega}_t)$ is the cosine factor and $V \in \{0,1\}$ is visibility. This solid-angle formulation directly motivates Rx-centric sampling: trace rays from the receiver, intersect the scene, and evaluate the integrand with a deterministic connection to the transmitter.

For multi-bounce propagation with path depth $D$ and interaction points $\mathbf{x}_1, \ldots, \mathbf{x}_D$, each stochastically sampled segment contributes a BSDF-over-PDF ratio and geometric coupling. Given $N$ Monte Carlo samples, the unbiased estimator is (derivation in the supplement, Sec.~\ref{sec:supp_mc_derivation}):
\begin{equation}
    \hat{P}_r^{(D)}
    =
    \frac{P_t\,\lambda^2}{(4\pi)^2}\,
    \frac{1}{N}\sum_{k=1}^{N}
    \frac{G_{r,k}}{\rho_{rx,k}}
    \prod_{\ell=1}^{D{-}1}
    \frac{f_\ell\, \mathcal{G}_\ell}{\rho_\ell}
    \;\cdot\;
    \frac{G_{t,k}\, c_{D{\to}tx}\, f_D}{d_{D,tx}^2}
    \, V_k,
    \label{eq:mc_multibounce}
\end{equation}
where $\mathcal{G}_\ell = c_{\ell{\to}\ell{+}1}\, c_{\ell{+}1{\to}\ell}\,/\,d_{\ell,\ell{+}1}^2$ is the geometric coupling between consecutive bounces, with $c_{\ell{\to}\ell{+}1} = \max(0, \mathbf{n}_\ell \cdot \boldsymbol{\omega}_{\ell{\to}\ell{+}1})$ the cosine factor at bounce $\ell$ toward bounce $\ell{+}1$ and $d_{\ell,\ell{+}1}$ the distance between them. The BSDF at each bounce is $f_\ell = f(\boldsymbol{\omega}_{\ell-1}, \mathbf{x}_\ell, \boldsymbol{\omega}_{\ell+1})$ (Sec.~\ref{sec:bsdf}), which decomposes each surface interaction into slab transmission, coherent specular reflection, incoherent diffuse scattering, and edge diffraction. The final term connects deterministically to the transmitter: $c_{D{\to}tx}$ is the cosine factor at $\mathbf{x}_D$ toward the Tx and $d_{D,tx}$ is their distance. The sampling PDFs $\rho_\ell$ are determined by the ray generation strategies in Sec.~\ref{sec:sampling}. The single-bounce case ($D{=}1$) reduces to $G_t\,G_r\,c_t\,f\,V\,/\,(d_t^2\,\rho_{rx})$, and we use Russian roulette~\cite{veach1998robust} for unbiased path termination.

The MC estimator provides a complex amplitude for each traced path. To synthesize raw FMCW ADC, each path $p$ for Tx $i$ and Rx $j$ contributes a phase-delayed chirp sample using the complex baseband form (Sec.~\ref{sec:supp_fmcw}--\ref{sec:supp_iq} in the supplement):
\begin{equation}
    s_{ij}^{(p)}[k]
    =
    a_{ij}^{(p)}
    \exp\!\Big(j\, 2\pi (f_c + S t_k)\,\tau_{ij}^{(p)} \Big),
    \label{eq:path_adc}
\end{equation}
where $k$ indexes fast-time ADC samples at times $t_k$, $j$ is the imaginary unit, $\tau_{ij}^{(p)} = R_{ij}^{(p)}/c$ is the round-trip delay for total path length $R_{ij}^{(p)}$ at speed of light $c$, $f_c$ is the carrier frequency, $S$ is the chirp slope, and $a_{ij}^{(p)} \in \mathbb{C}$ aggregates the MC throughput from Eq.~\eqref{eq:mc_multibounce}, namely antenna gains, BSDF evaluations, geometric coupling, and spreading losses. The rendered ADC sums coherently over all paths: $s_{ij}[k] = \sum_p s_{ij}^{(p)}[k]$.

Equations~\eqref{eq:radar_bistatic}--\eqref{eq:path_adc} describe a single Tx--Rx pair. In a MIMO array, each pair is evaluated independently through the same forward model. However, coherent MIMO imaging via FFT~\cite{9658500} requires exact relative phase across all pairs, so independent MC samples per receiver would corrupt this. Following the MIMO ray reuse strategy of Sch{\"u}{\ss}ler et al.~\cite{9533181}, we share path topology across all Tx--Rx pairs. A single set of interaction points is found from shared seeds, then evaluated for all pairs by vectorizing only the Tx$\rightarrow\mathbf{x}_1$ and $\mathbf{x}_D\rightarrow$Rx legs. Each pair produces unique amplitudes and phases but shares bounce topology, preserving MIMO coherence while reducing cost from $\mathcal{O}(N_\mathrm{Tx} \cdot N)$ to $\mathcal{O}(N)$.

\subsection{Physically-Based mmWave BSDF}
\label{sec:bsdf}

The BSDF $f$ appearing in the MC estimator (Eq.~\eqref{eq:mc_multibounce}) governs how electromagnetic energy is redistributed at each surface interaction. We model it using ITU-R P.2040 recommendations for mmWave propagation~\cite{itu_p2040}, inspired by the coherent spatially varying BSDF (CSVBSDF) formulation of Wei et al.~\cite{10679198}. Each mesh vertex stores six physically interpretable material parameters, namely real and imaginary relative permittivity $(\varepsilon_r', \varepsilon_r'')$, RMS surface roughness $\sigma_h$, correlation length $l_c$, a Kirchhoff-approximation / small-perturbation-method (KA/SPM) blend factor $\tau$, and slab thickness $d$. At each ray--triangle intersection, parameters are barycentric-interpolated from the three triangle vertices ($\mathbf{m}_i \in \mathbb{R}^6$), ensuring $C^0$-continuous material fields with gradients flowing to all three vertices.

The BSDF decomposes into coherent and incoherent components via the Rayleigh roughness factor~\cite{beckmann1987scattering} $\eta = \exp(-(2k\sigma_h\cos\theta_t)^2)$, $k=2\pi/\lambda$, where $\theta_t$ is the incidence angle between $\boldsymbol{\omega}_t$ and the surface normal:
\begin{equation}
    f = A(\boldsymbol{\omega}_t)\;\Big[\eta\,f_\mathrm{coh} + (1-\eta)\,f_\mathrm{inc}\Big],
    \label{eq:bsdf_full}
\end{equation}
where $A(\boldsymbol{\omega}_t) \in [0,1]$ is the slab Fresnel energy gate. The lobe decompositions are:
\begin{align}
    f_\mathrm{coh} &= \tau\, f_\mathrm{KA} + (1{-}\tau)\, f_\mathrm{SPM}, \label{eq:bsdf_coh} \\
    f_\mathrm{inc} &= \gamma\, f_\mathrm{dir} + (1{-}\gamma)\, f_\mathrm{broad}, \label{eq:bsdf_inc}
\end{align}
where $\tau$ is a learnable KA/SPM blend and $\gamma$ is a roughness-dependent sigmoid. The coherent lobes are $f_\mathrm{KA}$ and $f_\mathrm{SPM}$. $f_\mathrm{KA}$ is a GGX microfacet specular lobe~\cite{10.2312:EGWR/EGSR07/195-206} with roughness $\alpha = \sqrt{4\pi\sigma_h/\lambda}$, and $f_\mathrm{SPM}$ is a von Mises--Fisher (vMF) distribution centered on the specular direction with concentration $\kappa = \sqrt{l_c/\lambda}\,/\,(1 + 3\sigma_h/l_c)$~\cite{ulaby2015fundamentals}. The incoherent lobes are $f_\mathrm{dir}$, a broader vMF lobe, and $f_\mathrm{broad}$, Lambertian. The energy gate $A(\boldsymbol{\omega}_t)$ uses an ITU-R P.2040 slab model with thickness $d$ and complex permittivity $\varepsilon_r = \varepsilon_r' - j\varepsilon_r''$, capturing multiple internal reflections rather than single-interface Fresnel.

The Fresnel coefficients in the energy gate also determine polarization state, which we maintain throughout path tracing using a local $s/p$ basis. At each interaction, the complex Fresnel coefficients derived from learnable permittivity update both amplitude and phase of the $s$ and $p$ field components~\cite{born2013principles}. Between bounces, the polarization basis is transported deterministically to preserve phase consistency. The polarization-dependent energy gate is $A = f_s |r_s|^2 + f_p |r_p|^2$, where $r_s$, $r_p$ are the complex Fresnel reflection coefficients for the $s$ and $p$ polarizations and $f_s$, $f_p$ are the antenna-dependent power fractions in each polarization.

\subsection{Ray Generation}
\label{sec:sampling}

The MC estimator (Eq.~\eqref{eq:mc_multibounce}) requires sampling ray directions at each interaction. Because the mmWave BSDF (Sec.~\ref{sec:bsdf}) combines smooth diffuse, near-delta specular, and edge-diffraction components that no single sampling PDF can efficiently cover, we provide the PDFs $\rho_\ell$ via three complementary strategies, multiple importance sampling (MIS)~\cite{10.1145/218380.218498} and balance heuristic weights.

\paragraph{Rx-centric reservoir sampling.}
\label{sec:surface_sampling}
Because mmWave antennas are highly directional, we employ Rx-centric reservoir sampling inspired by Resampled Importance Sampling (RIS)~\cite{10.2312:EGWR/EGSR05/139-146} and spatiotemporal reservoir resampling (ReSTIR)~\cite{10.1145/3386569.3392481, 10.1145/3528223.3530158}. For each Rx element, we sample $M$ candidate directions from a cosine-weighted hemisphere proposal centered on the Rx boresight, $\rho_{rx}(\boldsymbol{\omega}) = \cos\theta / \pi$, which naturally concentrates samples within the antenna main lobe, then retain $K$ directions that successfully intersect scene geometry.

\paragraph{Specular manifold sampling.}
\label{sec:sms}
LiDAR-derived meshes contain triangles much smaller than the mmWave wavelength ($\lambda \approx 3.9$~mm). The image method (mirroring Tx across the hit triangle's plane) frequently fails because the exact specular point falls outside the original triangle. We instead use Specular Manifold Sampling (SMS)~\cite{10.1145/3386569.3392408}, which solves for exact reflection points via Newton iteration on the half-vector constraint $\mathbf{C}(\mathbf{x}) = [\mathbf{s}{\cdot}\mathbf{h},\; \mathbf{t}{\cdot}\mathbf{h}]^\top = \mathbf{0}$, where $\mathbf{s},\mathbf{t}$ are surface tangents and $\mathbf{h}$ is the half-vector. After each Newton step, re-projection onto the mesh via ray tracing allows the solver to cross triangle boundaries, producing exact specular paths with deterministic Fresnel weights.

We extend SMS for radar inverse rendering with GPU-vectorized Newton iteration via DrJit~\cite{10.1145/3528223.3530099}, Implicit Function Theorem gradient support that stores the Jacobian inverse at convergence for backpropagation through vertex positions, and multi-bounce specular chain solving via a block-tridiagonal algorithm.

\paragraph{Free-space diffraction BSDF.}
\label{sec:fsd}
Unlike the Uniform Theory of Diffraction (UTD), which requires explicit edge detection and separate ray classes, the free-space diffraction (FSD) BSDF~\cite{10.1145/3658166} formulates edge diffraction as a standard scattering distribution compatible with path tracing. At each intersection, the FSD BSDF projects nearby triangles onto a virtual screen, identifies boundary edges, and evaluates closed-form Fraunhofer diffraction integrals. A fraction $\beta$ of reflected energy is redirected into importance-sampleable diffracted lobes.

We extend FSD for mmWave radar with: (i) material-dependent edge opacity via Fresnel reflectance from learnable permittivity; (ii) Jones polarization tracking through diffracted paths for MIMO phase coherence; and (iii) edge suppression filters that reject artifacts from noisy LiDAR tessellation.

\subsection{Differentiable Optimization}
\label{sec:optimization}

We optimize per-vertex ITU physics materials ($6 \times N_v$ parameters with sigmoid/exponential reparameterizations for physical bounds), per-vertex normals $\mathbf{N} \in \mathbb{R}^{N_v \times 3}$, and antenna beam patterns. The pipeline additionally supports differentiable vertex positions and 6-DoF radar pose, but we keep these fixed in our experiments, since our coarse-to-fine LiDAR-radar alignment (Sec.~\ref{sec:supp_alignment} in the supplement) already provides sub-centimeter translational and sub-degree rotational accuracy, so the remaining sim-to-real gap is dominated by material and scattering properties rather than geometric misregistration. Vertex-position and pose optimization remain available as a fallback when the input mesh is degraded or alignment is poor (Sec.~\ref{sec:supp_training}, and limitations in Sec.~\ref{sec:limitations}). Initialization uses ITU-R P.2040 reference values for common materials (concrete, glass, metal). Details on bounds, learning rates, and schedules are in the supplement (Sec.~\ref{sec:supp_training}).

The entire forward model (Secs.~\ref{sec:forward_model}--\ref{sec:sampling}) executes inside a single DrJit~\cite{10.1145/3528223.3530099} computation graph with automatic differentiation. Gradients flow from $\mathcal{L}$ through the ADC, per-path phasors, BSDF evaluations, and ray--triangle intersections back to all learnable parameters in a single backward pass. At silhouette edges where visibility changes discontinuously, projective boundary gradient sampling~\cite{10.1145/3618385} corrects gradient bias for vertex positions and pose when these are enabled.

\paragraph{Loss function.}
We minimize a range--azimuth reconstruction loss on min--max normalized magnitudes:
\begin{equation}
    \mathcal{L}
    =
    \frac{1}{|\Gamma|}
    \sum_{(r,a)\in\Gamma}
    \big\lVert \widetilde{\mathrm{RA}}_\mathrm{render}(r,a) - \widetilde{\mathrm{RA}}_\mathrm{gt}(r,a) \big\rVert_2^2,
    \label{eq:ra_loss}
\end{equation}
where $\Gamma$ is the set of range--azimuth bins, $\widetilde{\mathrm{RA}} = (\mathrm{RA} - \min\mathrm{RA}) / (\max\mathrm{RA} - \min\mathrm{RA})$ denotes min--max normalization, and RA maps are obtained via the TDM-MIMO FFT pipeline~\cite{9658500}. RA magnitude validates phase coherence because the azimuth FFT encodes both relative phase across virtual elements and ADC magnitude. This is ideal, because the absolute phase depends on user-configurable initial phase, oscillator drift, and sub-millimeter path-length variation at 77\,GHz and carries no reliable supervisory signal (Sec.~\ref{sec:phase_sensitivity_loss_landscape}). After convergence, we freeze optimized parameters and re-render from dense virtual apertures to synthesize 3D radar data (Sec.~\ref{sec:3d_occupancy}).

\section{Experiments}

\begin{figure}[t]
  \centering
  \includegraphics[width=\linewidth]{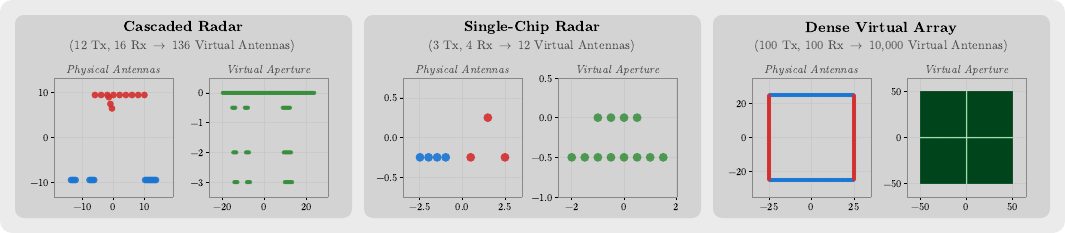}
  \caption{Antenna configurations: cascaded MMWCAS (12Tx$\times$16Rx with 136 unique virtual elements) used for training, single-chip AWR1843 (3Tx$\times$4Rx with 12 unique virtual elements) used for transfer evaluation, dense array (100Tx$\times$100Rx, 10{,}000 virtual elements) used for 3D inference. Top: physical Tx/Rx; bottom: virtual aperture. The cascade's near-1D virtual aperture yields $256{\times}127{\times}1$ RAE resolution (Sec.~\ref{sec:supp_fmcw},~\ref{sec:tdm_mimo_imaging}); only the dense array resolves true elevation at $256{\times}127{\times}127$ RAE resolution.}
  \label{fig:antenna_layouts}
\end{figure}

We evaluate mmIR on seven diverse outdoor scenes from the ColoRadar dataset~\cite{doi:10.1177/02783649211068535}, which provides synchronized mmWave radar and LiDAR captures with raw ADC access. We design three experiments that progressively test our forward model under increasingly challenging conditions: (1)~training evaluation on the cascaded imaging radar used for optimization (12Tx$\times$16Rx with 136 unique virtual elements, spanning 86 uniformly sampled virtual elements along the azimuth); (2)~cross-sensor transfer to a co-located single-chip radar (3Tx$\times$4Rx with 12 virtual elements, spanning 8 uniformly sampled virtual elements along the azimuth) without re-training; and (3)~3D occupancy inference via a dense virtual array (100Tx$\times$100Rx with 10{,}000 unique virtual elements, spanning 100 uniformly sampled virtual elements along the azimuth and the elevation each). Figure~\ref{fig:antenna_layouts} illustrates all three radar configurations and antenna layouts. Dataset details, scene geometry reconstruction, and implementation hyperparameters are provided in the supplement (Sec.~\ref{sec:supp_implementation}).

\subsection{Training Evaluation: ADC Rendering and Range--Azimuth}
\label{sec:training_ra}

\begin{table}[t]
\centering
\caption{Range--azimuth (RA) and ADC training evaluation on seven outdoor and six indoor scenes. mmIR is compared against Sionna-RT~\cite{10465179}. Metrics: Pearson correlation (Corr), PSNR, SSIM, RMSE, and ADC-level magnitude error. Best in \textbf{bold}. Results use the min--max-normalized RA loss (Eq.~\ref{eq:ra_loss}).}
\label{tab:training_ra}
\resizebox{\linewidth}{!}{%
\begin{tabular}{l cc cc cc cc cc cc}
\toprule
& \multicolumn{2}{c}{RA Correlation $\uparrow$} & \multicolumn{2}{c}{RA PSNR $\uparrow$} & \multicolumn{2}{c}{RA SSIM $\uparrow$} & \multicolumn{2}{c}{RA RMSE $\downarrow$} & \multicolumn{2}{c}{ADC Log-Mag MSE $\downarrow$} & \multicolumn{2}{c}{ADC MM-Mag MSE $\downarrow$} \\
\cmidrule(lr){2-3} \cmidrule(lr){4-5} \cmidrule(lr){6-7} \cmidrule(lr){8-9} \cmidrule(lr){10-11} \cmidrule(lr){12-13}
Scene & Ours & Sionna-RT & Ours & Sionna-RT & Ours & Sionna-RT & Ours & Sionna-RT & Ours & Sionna-RT & Ours & Sionna-RT \\
\midrule
\multicolumn{13}{l}{\textit{Outdoor (7 scenes)}} \\
  S0-F135 & \textbf{0.921} & 0.418 & \textbf{37.4} & 21.4 & \textbf{0.938} & 0.509 & \textbf{0.0135} & 0.0850 & \textbf{48.3271} & 244.8361 & \textbf{0.0421} & 0.0469 \\
  S0-F390 & \textbf{0.936} & 0.242 & \textbf{41.0} & 23.4 & \textbf{0.937} & 0.550 & \textbf{0.0089} & 0.0674 & \textbf{5.0250} & 294.3032 & \textbf{0.0289} & 0.0417 \\
  S1-F185 & \textbf{0.954} & 0.312 & \textbf{39.8} & 18.9 & \textbf{0.926} & 0.326 & \textbf{0.0102} & 0.1138 & \textbf{32.3385} & 282.6453 & \textbf{0.0398} & 0.0461 \\
  S1-F438 & \textbf{0.942} & 0.346 & \textbf{37.7} & 19.9 & \textbf{0.926} & 0.401 & \textbf{0.0130} & 0.1008 & \textbf{32.5274} & 288.9782 & \textbf{0.0369} & 0.0435 \\
  S2-F105 & \textbf{0.887} & 0.188 & \textbf{34.6} & 28.7 & \textbf{0.894} & 0.724 & \textbf{0.0185} & 0.0367 & \textbf{23.5161} & 244.9075 & 0.0498 & \textbf{0.0481} \\
  S2-F160 & \textbf{0.864} & 0.370 & \textbf{34.0} & 22.4 & \textbf{0.789} & 0.519 & \textbf{0.0200} & 0.0763 & \textbf{27.6683} & 275.3327 & \textbf{0.0444} & 0.0461 \\
  S2-F300 & \textbf{0.930} & 0.388 & \textbf{38.7} & 21.3 & \textbf{0.906} & 0.512 & \textbf{0.0117} & 0.0858 & \textbf{30.0014} & 247.1371 & \textbf{0.0370} & 0.0543 \\
  Outdoor Mean & \textbf{0.919$\pm$0.030} & 0.324$\pm$0.076 & \textbf{37.6$\pm$2.4} & 22.3$\pm$3.0 & \textbf{0.902$\pm$0.049} & 0.506$\pm$0.115 & \textbf{0.0137$\pm$0.0038} & 0.0808$\pm$0.0229 & \textbf{28.4863$\pm$11.9698} & 268.3057$\pm$20.3703 & \textbf{0.0399$\pm$0.0061} & 0.0467$\pm$0.0037 \\
\midrule
\multicolumn{13}{l}{\textit{Indoor (6 scenes)}} \\
  S3-F453 & \textbf{0.839} & 0.168 & \textbf{36.5} & 28.2 & \textbf{0.860} & 0.724 & \textbf{0.0150} & 0.0391 & \textbf{26.1754} & 304.6365 & 0.0418 & \textbf{0.0389} \\
  S4-F353 & \textbf{0.970} & 0.366 & \textbf{45.8} & 18.3 & \textbf{0.936} & 0.485 & \textbf{0.0051} & 0.1220 & \textbf{27.5573} & 262.6228 & 0.0489 & \textbf{0.0440} \\
  S5-F497 & \textbf{0.984} & 0.371 & \textbf{52.8} & 27.1 & \textbf{0.983} & 0.864 & \textbf{0.0023} & 0.0441 & \textbf{32.9267} & 272.8659 & \textbf{0.0459} & 0.0508 \\
  S5-F527 & \textbf{0.791} & 0.111 & \textbf{39.6} & 25.4 & \textbf{0.915} & 0.848 & \textbf{0.0105} & 0.0538 & \textbf{6.0017} & 293.5447 & \textbf{0.0431} & 0.0474 \\
  S6-F175 & \textbf{0.945} & 0.342 & \textbf{40.3} & 27.5 & \textbf{0.897} & 0.589 & \textbf{0.0096} & 0.0420 & \textbf{37.0109} & 265.2215 & \textbf{0.0378} & 0.0534 \\
  S6-F575 & \textbf{0.924} & 0.372 & \textbf{42.2} & 20.6 & \textbf{0.917} & 0.569 & \textbf{0.0077} & 0.0931 & \textbf{15.7710} & 269.3423 & \textbf{0.0368} & 0.0391 \\
  Indoor Mean & \textbf{0.909$\pm$0.070} & 0.288$\pm$0.107 & \textbf{42.9$\pm$5.2} & 24.5$\pm$3.7 & \textbf{0.918$\pm$0.037} & 0.680$\pm$0.143 & \textbf{0.0084$\pm$0.0040} & 0.0657$\pm$0.0311 & \textbf{24.2405$\pm$10.4695} & 278.0390$\pm$15.5574 & \textbf{0.0424$\pm$0.0042} & 0.0456$\pm$0.0055 \\
\midrule
  Overall Mean & \textbf{0.914$\pm$0.053} & 0.307$\pm$0.093 & \textbf{40.0$\pm$4.8} & 23.3$\pm$3.5 & \textbf{0.910$\pm$0.044} & 0.586$\pm$0.155 & \textbf{0.0112$\pm$0.0047} & 0.0738$\pm$0.0280 & \textbf{26.5267$\pm$11.4986} & 272.7980$\pm$18.9390 & \textbf{0.0410$\pm$0.0055} & 0.0462$\pm$0.0046 \\
\bottomrule
\end{tabular}
}
\end{table}

\begin{figure}[t]
  \centering
  \includegraphics[width=\linewidth]{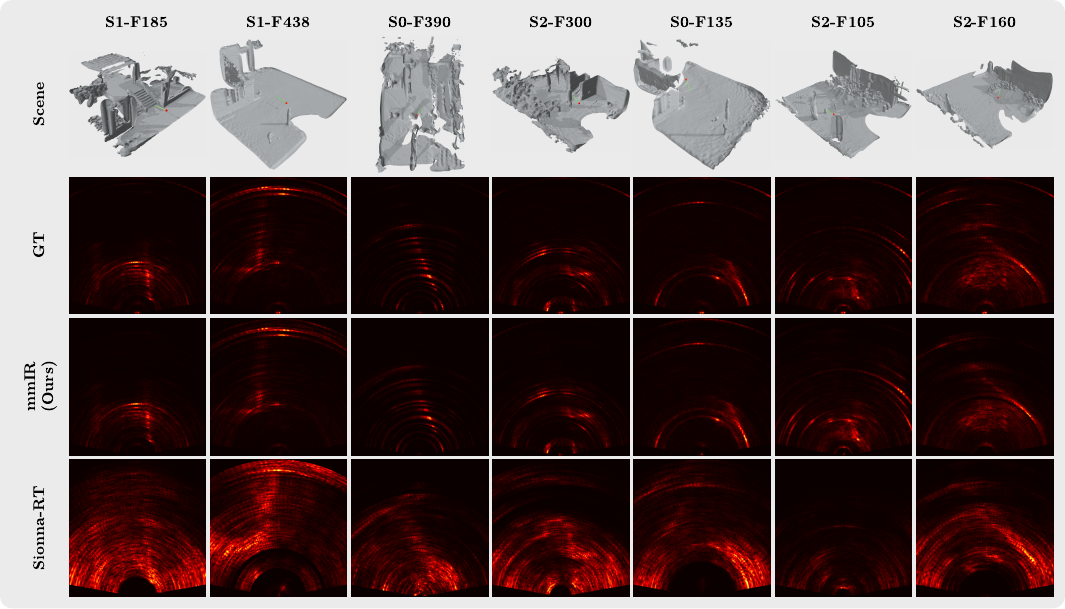}
  \caption{Qualitative RA on the seven outdoor scenes, sorted by mmIR correlation. Rows: scene mesh (red sphere\,=\,radar, green arrow\,=\,boresight), GT, mmIR, Sionna-RT. mmIR captures dominant scatterers and multipath; Sionna-RT is weaker and misaligned.}
  \label{fig:eval_ra}
\end{figure}

\begin{figure}[t]
  \centering
  \includegraphics[width=\linewidth]{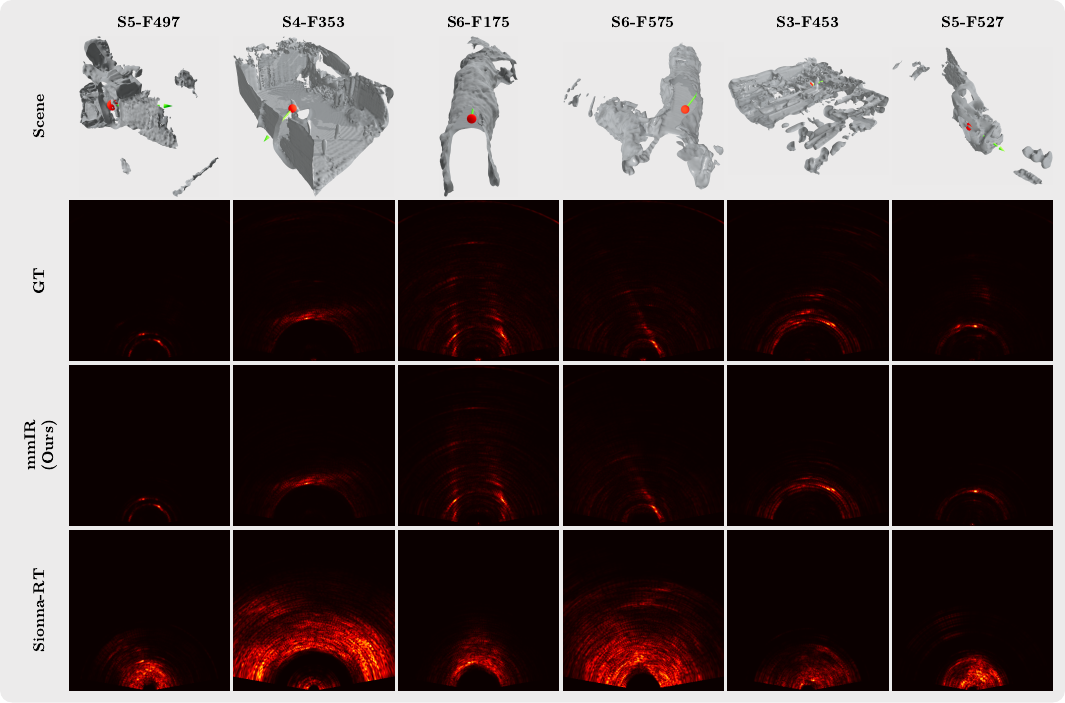}
  \caption{As Fig.~\ref{fig:eval_ra}, for six indoor scenes (lab, hallway, classroom). mmIR tracks the GT returns; Sionna-RT over-responds in dense indoor multipath.}
  \label{fig:eval_ra_indoor}
\end{figure}

We compare rendered and measured range--azimuth (RA) maps on the cascaded radar used for optimization. Table~\ref{tab:training_ra} reports Pearson correlation, PSNR, SSIM, RMSE, and ADC-level magnitude metrics across all seven scenes, with Sionna-RT~\cite{10465179} as baseline. Just as the loss is defined on RA magnitude (Eq.~\ref{eq:ra_loss}), we evaluate in the RA domain because it implicitly validates phase accuracy. The azimuth FFT converts relative phase differences across virtual elements into angular information, so any systematic phase error due to per-element bias, incorrect path-length scaling, or incoherent phasor accumulation would smear or shift targets in the angular domain, degrading all RA metrics. Absolute phase is not meaningful in practice, as it depends on user-configurable initial phase, oscillator drift, and phase noise. These common-mode offsets appear in the DC term of the spatial FFT and are dropped during processing (Sec.~\ref{sec:phase_sensitivity_loss_landscape}).

mmIR achieves a mean Pearson correlation of \textbf{0.919} on the seven outdoor scenes and \textbf{0.909} on six additional ColoRadar indoor scenes (lab, hallway, classroom), for an overall mean of \textbf{0.914}, substantially outperforming Sionna-RT (0.324/0.288/0.307 outdoor/indoor/overall). Table~\ref{tab:training_ra} quantifies this gap across all metrics, and Figs.~\ref{fig:eval_ra} and~\ref{fig:eval_ra_indoor} provide the qualitative comparisons, sorted by correlation. mmIR reproduces both the dominant scatterer locations and the diffuse multipath structure present in the ground truth, whereas Sionna-RT produces attenuated responses with spatially misaligned features.

Three architectural differences drive this gap. First, mmIR's physically-based BSDF with six per-vertex learnable parameters captures spatially varying reflectance that Sionna-RT's per-object scattering model cannot represent. This is most evident in geometrically complex scenes such as S1-F438 and S0-F135, where mixed-material surfaces dominate the radar response. Second, mmIR jointly optimizes surface normals and antenna beam patterns, parameters entirely outside Sionna-RT's optimization scope, correcting both surface orientation errors inherited from noisy LiDAR tessellation and systematic antenna gain biases. Third, Sionna-RT's path solver executes outside the AD graph. Ray tracing runs under gradient suspension and all path geometry is subsequently detached, so while material gradients propagate through field coefficients at each bounce, the path topology itself is frozen and no gradients reach vertex positions or normals, and a material change at one surface cannot redirect subsequent bounces. mmIR places the entire forward model inside a single computation graph, enabling end-to-end gradient flow from the loss through ADC synthesis, phasor accumulation, BSDF evaluation, and ray--triangle intersections. The weakest outdoor scene, S2-F160 (0.864), shows that even challenging geometries with complex multipath achieve high correlation under the min--max-normalized loss; residual gaps are dominated by sub-wavelength LiDAR-to-radar misalignment (Sec.~\ref{sec:supp_alignment}), to which coherent radar rendering is sensitive (Sec.~\ref{sec:limitations}). mmIR substantially outperforms the baseline on every scene, indicating that the richer parameter space and deeper gradient flow consistently improve fit quality.

\subsection{Radar Transfer Evaluation}
\label{sec:radar_transfer}

\begin{table}[t]
\centering
\caption{Cross-sensor transfer evaluation. Parameters trained on the cascade (12Tx$\times$16Rx) are frozen and used to render the single-chip AWR1843 (3Tx$\times$4Rx) without re-training. Best in \textbf{bold}.}
\label{tab:radar_transfer}
\resizebox{\linewidth}{!}{%
\begin{tabular}{l cc cc cc cc cc cc}
\toprule
& \multicolumn{2}{c}{RA Correlation $\uparrow$} & \multicolumn{2}{c}{RA PSNR $\uparrow$} & \multicolumn{2}{c}{RA SSIM $\uparrow$} & \multicolumn{2}{c}{RA RMSE $\downarrow$} & \multicolumn{2}{c}{ADC Log-Mag MSE $\downarrow$} & \multicolumn{2}{c}{ADC MM-Mag MSE $\downarrow$} \\
\cmidrule(lr){2-3} \cmidrule(lr){4-5} \cmidrule(lr){6-7} \cmidrule(lr){8-9} \cmidrule(lr){10-11} \cmidrule(lr){12-13}
Scene & Ours & Bench. & Ours & Bench. & Ours & Bench. & Ours & Bench. & Ours & Bench. & Ours & Bench. \\
\midrule
  S0-F135 & \textbf{0.533} & 0.002 & \textbf{25.6} & 17.8 & \textbf{0.505} & 0.464 & \textbf{0.0524} & 0.1283 & \textbf{10.9639} & 301.4341 & 0.2194 & \textbf{0.0668} \\
  S0-F390 & \textbf{0.464} & 0.328 & 21.2 & \textbf{24.2} & 0.529 & \textbf{0.566} & 0.0873 & \textbf{0.0619} & \textbf{12.9644} & 331.2549 & \textbf{0.0630} & 0.0840 \\
  S1-F185 & \textbf{0.567} & 0.136 & \textbf{22.4} & 20.1 & 0.422 & \textbf{0.480} & \textbf{0.0759} & 0.0990 & \textbf{0.5754} & 324.8441 & 0.1993 & \textbf{0.0771} \\
  S1-F438 & \textbf{0.684} & 0.079 & \textbf{23.2} & 13.5 & \textbf{0.674} & 0.202 & \textbf{0.0694} & 0.2119 & \textbf{21.8672} & 348.0237 & 0.1507 & \textbf{0.0662} \\
  S2-F105 & \textbf{0.608} & 0.084 & \textbf{24.7} & 22.8 & 0.609 & \textbf{0.690} & \textbf{0.0583} & 0.0721 & \textbf{3.4458} & 289.7929 & 0.1959 & \textbf{0.1068} \\
  S2-F160 & \textbf{0.276} & 0.159 & \textbf{19.3} & 18.3 & \textbf{0.493} & 0.463 & \textbf{0.1085} & 0.1219 & \textbf{20.1530} & 338.1699 & 0.0893 & \textbf{0.0798} \\
  S2-F300 & \textbf{0.745} & 0.008 & \textbf{27.2} & 23.6 & \textbf{0.468} & 0.412 & \textbf{0.0434} & 0.0661 & \textbf{7.7226} & 287.9379 & 0.1074 & \textbf{0.0948} \\
\midrule
  Mean & \textbf{0.554$\pm$0.143} & 0.114$\pm$0.103 & \textbf{23.4$\pm$2.5} & 20.0$\pm$3.6 & \textbf{0.529$\pm$0.080} & 0.468$\pm$0.138 & \textbf{0.0707$\pm$0.0206} & 0.1087$\pm$0.0488 & \textbf{11.0989$\pm$7.3886} & 317.3511$\pm$22.3659 & 0.1464$\pm$0.0565 & \textbf{0.0822$\pm$0.0136} \\
\bottomrule
\end{tabular}
}
\end{table}

A central claim of physics-based inverse rendering is that the optimized scene captures true physical properties, not sensor-specific patterns. We validate this by \emph{cross-sensor transfer}: the scene parameters trained on the cascaded radar (12Tx$\times$16Rx spanning 86 uniformly sampled virtual elements along azimuth) are frozen and used to render ADC for the single-chip radar (3Tx$\times$4Rx, 12 virtual elements total spanning 8 uniformly sampled virtual elements along azimuth) that shares the same physical mounting. We apply the same rigid-body alignment transformation learned during cascaded training to the single-chip configuration and render forward-only. In Table~\ref{tab:radar_transfer}, mmIR achieves a mean transfer correlation of \textbf{0.554} on outdoor scenes, outperforming Sionna-RT (0.114). This indicates that mmIR has learned transferable scene physics rather than memorizing the training sensor's idiosyncrasies, via optimized materials and normals that generalize across sensor configurations.

\subsection{Inference: 3D Occupancy via Dense Virtual Array}
\label{sec:3d_occupancy}

\begin{table}[t]
\centering
\caption{3D occupancy vs.\ GT LiDAR. Baseline: ColoRadar's cascaded-radar point cloud with elevation extrusion added in post-processing. Ours: mmIR's RAE rendered via the dense virtual aperture (100$\times$100 elements), resolving true elevation. KD-tree match at $\tau{=}0.5$\,m. Best in \textbf{bold}.}
\label{tab:3d_occupancy}
\resizebox{\linewidth}{!}{%
\begin{tabular}{l cc cc cc cc cc cc}
\toprule
& \multicolumn{2}{c}{Precision $\uparrow$} & \multicolumn{2}{c}{Recall $\uparrow$} & \multicolumn{2}{c}{F1 $\uparrow$} & \multicolumn{2}{c}{Accuracy $\uparrow$} & \multicolumn{2}{c}{RMSE $\downarrow$} & \multicolumn{2}{c}{R-CD $\downarrow$} \\
\cmidrule(lr){2-3} \cmidrule(lr){4-5} \cmidrule(lr){6-7} \cmidrule(lr){8-9} \cmidrule(lr){10-11} \cmidrule(lr){12-13}
Scene & Ours & ColoR & Ours & ColoR & Ours & ColoR & Ours & ColoR & Ours & ColoR & Ours & ColoR \\
\midrule
  S0-F135 & 0.077 & \textbf{0.223} & 0.018 & \textbf{0.054} & 0.029 & \textbf{0.087} & 0.044 & \textbf{0.057} & \textbf{2.324} & 3.302 & 0.248 & \textbf{0.228} \\
  S0-F390 & \textbf{0.481} & 0.045 & \textbf{0.110} & 0.014 & \textbf{0.179} & 0.021 & \textbf{0.212} & 0.017 & \textbf{2.692} & 2.745 & 0.216 & \textbf{0.207} \\
  S1-F185 & \textbf{0.997} & 0.483 & \textbf{0.566} & 0.082 & \textbf{0.722} & 0.141 & \textbf{0.703} & 0.103 & \textbf{0.909} & 2.380 & \textbf{0.067} & 0.236 \\
  S1-F438 & \textbf{0.633} & 0.433 & \textbf{0.265} & 0.117 & \textbf{0.373} & 0.185 & \textbf{0.362} & 0.131 & \textbf{1.511} & 3.339 & \textbf{0.116} & 0.268 \\
  S2-F105 & \textbf{0.771} & 0.154 & \textbf{0.162} & 0.025 & \textbf{0.268} & 0.044 & \textbf{0.340} & 0.029 & \textbf{1.099} & 3.402 & \textbf{0.113} & 0.282 \\
  S2-F160 & \textbf{0.516} & 0.383 & \textbf{0.177} & 0.165 & \textbf{0.264} & 0.231 & \textbf{0.257} & 0.183 & 1.917 & \textbf{1.611} & 0.146 & \textbf{0.118} \\
  S2-F300 & \textbf{0.850} & 0.508 & \textbf{0.501} & 0.122 & \textbf{0.630} & 0.197 & \textbf{0.601} & 0.130 & \textbf{1.040} & 1.643 & \textbf{0.083} & 0.118 \\
\midrule
  Mean & \textbf{0.618$\pm$0.278} & 0.319$\pm$0.165 & \textbf{0.257$\pm$0.189} & 0.083$\pm$0.051 & \textbf{0.352$\pm$0.228} & 0.129$\pm$0.075 & \textbf{0.360$\pm$0.210} & 0.093$\pm$0.056 & \textbf{1.642$\pm$0.638} & 2.632$\pm$0.721 & \textbf{0.141$\pm$0.062} & 0.208$\pm$0.061 \\
\bottomrule
\end{tabular}
}
\end{table}

\begin{figure}[t]
  \centering
  \includegraphics[width=\linewidth]{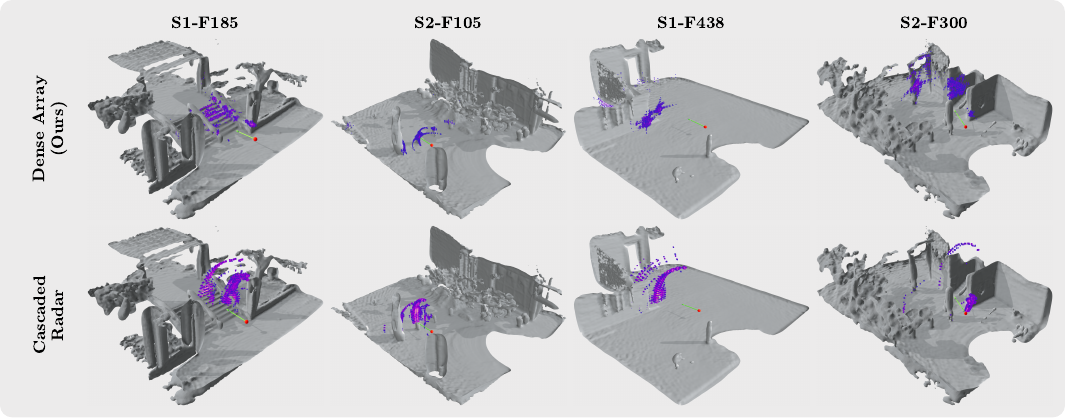}
  \caption{Single-frame 3D occupancy on the LiDAR mesh. Top: mmIR's rendered RAE via the dense virtual aperture (100$\times$100 elements), thresholded at the 99.9th intensity percentile. Bottom: ColoRadar-provided RAE, obtained from the near-1D cascaded radar with elevation extrusion added in post-processing. Red sphere\,=\,radar, green arrow\,=\,boresight.}
  \label{fig:3d_single_frame_comparison}
\end{figure}

After training, we freeze the optimized scene parameters and re-render from a dense virtual array (100Tx$\times$100Rx, 10{,}000 virtual elements with $\lambda/2$ spacing) inspired by the Rohde \& Schwarz QAR50~\cite{10.1145/3796224}. The cascade's near-1D virtual aperture yields $256{\times}127{\times}1$ RAE resolution (Sec.~\ref{sec:supp_fmcw},~\ref{sec:tdm_mimo_imaging}); only the dense array resolves true elevation at $256{\times}127{\times}127$ RAE resolution. We convert synthesized ADC into a 3D tensor via Range--Azimuth--Elevation (RAE) FFT and extract a sparse radar point cloud by thresholding at the 99.9th intensity percentile (Fig.~\ref{fig:3d_single_frame_comparison}). Following~\cite{11444848}, we use single-bounce propagation at test time for the dense array.

We evaluate against LiDAR after radar-guided filtering: (1) filtering by the dense array's 0~dB beam limits, (2) projecting LiDAR points onto the radar polar voxel grid, and (3) retaining only points where radar normalized intensity $\geq 0.1$. Occupancy metrics use KD-tree nearest neighbors with $\tau=0.5$~m threshold. Geometric fidelity is measured via RMSE and Relative Chamfer Distance (R-CD). The geometric input is a smoothed Poisson mesh aggregated from ${\sim}50$ LiDAR sweeps, the evaluation ground truth is a single raw LiDAR frame at a different time instant, and mmIR's output is the radar-physics-gated $99.9$th-percentile RAE peak set, not the input mesh (Sec.~\ref{sec:supp_postprocess}). Table~\ref{tab:3d_occupancy} shows substantial variation across outdoor scenes driven by radar visibility. Scenes with multiple surfaces within the field-of-view achieve strong reconstruction (S1-F185: Precision=0.997, R-CD=0.067), while open scenes with ground planes nearly parallel to boresight show degraded recall as specular energy is directed away from receivers.
Component ablations (Sec.~\ref{sec:ablation}) confirm each module contributes (e.g., $-3.7\%$ correlation without antenna optimization, $-1.0\%$ without diffraction). The supplement details training and implementation hyperparameters (Sec.~\ref{sec:supp_training}), runtime (Sec.~\ref{sec:runtime}), dense-array inference and 3D-occupancy post-processing (Sec.~\ref{sec:supp_inference},~\ref{sec:supp_postprocess}), and additional qualitative results and ablations (Sec.~\ref{sec:supp_qualitative}).

\section{Conclusion}

We introduced mmIR, an end-to-end differentiable inverse renderer that fits per-vertex ITU materials, normals, and antenna patterns from raw FMCW ADC via automatic differentiation through a phase-coherent MIMO forward model. On seven outdoor and six indoor scenes, mmIR achieves 0.914 mean Pearson correlation versus 0.307 for Sionna-RT on rendered RA images. We further validate cross-sensor generalization by transferring trained scenes to a different radar configuration. We then show that mmIR's fitted scenes re-render dense virtual-aperture ADC for 3D occupancy detection and scene reconstruction, validated against LiDAR. This opens a path toward physics-grounded radar sensing pipelines, including calibration, radar-aware 3D reconstruction, and joint multi-modal optimization~\cite{10.1145/3528223.3530161}.

\textbf{Limitations and future work.} mmIR's rendered ADC is bottlenecked by mesh quality, and the inverse renderer is sensitive to LiDAR-radar alignment for both cascade and single-chip configurations. Our method supports surface-only interactions and single-bounce diffraction, and does not model transmission, refraction, or volumetric scattering. We do not account for Doppler phase from ego-motion, which would require a TDM-compatible $4\pi\mathbf{v}_\mathrm{ego}\!\cdot\!t/\lambda$ term, and assume static scenes, leaving dynamic-scene modeling for future work. Vertex-position and pose gradients are supported, but we found that at 77~GHz ($\lambda \approx 3.9$~mm) sub-millimeter shifts induce large phase changes that destabilize these gradients under current Monte Carlo budgets and are therefore disabled. True shape-from-radar remains compelling but will likely require dense multi-view supervision, regularization strategies and priors beyond our single-frame RA fitting setting. Evaluation is also confined to ColoRadar's clear-weather captures, with adverse weather evaluation left to future work. Validating the inferred materials against ground-truth dielectric properties, for instance through controlled anechoic-chamber captures of known materials, is another promising direction, since ColoRadar provides no co-located RGB or ground-truth materials for this comparison. See the supplement (Sec.~\ref{sec:limitations}) for details.

\bibliographystyle{splncs04}
\bibliography{main}

\end{document}